\documentclass[10pt,letterpaper,twocolumn]{article}
\usepackage[latin9]{inputenc}
\usepackage{color}
\usepackage{array}
\usepackage{textcomp}
\usepackage{multirow}
\usepackage{amsmath}
\usepackage{amssymb}
\usepackage{graphicx}
\usepackage[unicode=true,pdfusetitle,
 bookmarks=false,
 breaklinks=true,pdfborder={0 0 1},backref=section,colorlinks=false]
 {hyperref}
\hypersetup{
 pagebackref=true,letterpaper=true,colorlinks}

\makeatletter

\pdfpageheight\paperheight
\pdfpagewidth\paperwidth

\providecommand{\tabularnewline}{\\}
\newcommand{\lyxdot}{.}

\usepackage{cvpr}\usepackage{times}\usepackage{epsfig}
\usepackage{xcolor}

 \cvprfinalcopy 

\def\cvprPaperID{1605} 

\ifcvprfinal\fi

\@ifundefined{showcaptionsetup}{}{%
 \PassOptionsToPackage{caption=false}{subfig}}
\usepackage{subfig}
\makeatother

\begin{document}

\title{Hand-Object Interaction and Precise Localization in Transitive Action Recognition}

\author{Amir Rosenfeld and Shimon Ullman\\  Department of Computer Science \& Applied Mathematics\\  Weizmann Institute of Science   \\  Rehovot, Israel \\ {\tt \{amir.rosenfeld,shimon.ullman\}@weizmann.ac.il}}
\maketitle
\begin{abstract}
Abstract Action recognition in still images has seen major improvement in recent years due to advances in human pose estimation, object recognition and stronger feature representations produced by deep neural networks. However, there are still many cases in which performance remains far from that of humans. A major difficulty arises in distinguishing between transitive actions in which the overall actor pose is similar, and recognition therefore depends on details of the grasp and the object, which may be largely occluded. In this paper we demonstrate how recognition is improved by obtaining precise localization of the action-object and consequently extracting details of the object shape together with the actor-object interaction. To obtain exact localization of the action object and its interaction with the actor, we employ a coarse-to-fine approach which combines semantic segmentation and contextual features, in successive stages. We focus on (but are not limited) to face-related actions, a set of actions that includes several currently challenging categories. We present an average relative improvement of 35\% over state-of-the art and validate through experimentation the effectiveness of our approach.
\end{abstract}

\section{Introduction}

\begin{figure}
\includegraphics[width=0.5\columnwidth]{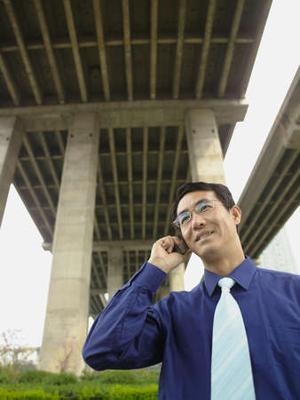}\includegraphics[width=0.5\columnwidth]{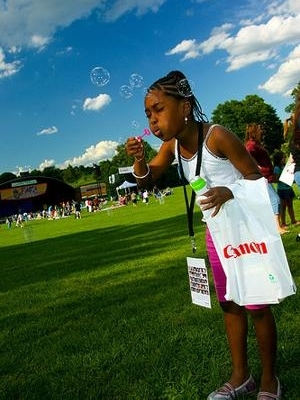}

\protect\caption{\label{fig:Actions-with-small}Actions where objects are barely visible.
Accurately localizing the action-object is important for succeeding
in classifying such images.}
\end{figure}

Recognizing actions in still images has been an active field of research
in recent years, using multiple benchmarks \cite{delaitre2010recognizing,pascal-voc-2012,sadeghi2011recognition,ICCV11_0744}.
It is a challenging problem since it combines different aspects of
recognition, including object recognition, human pose estimation,
and human-object interaction. Action recognition schemes can be divided
into transitive vs. intransitive. Our focus is on the recognition
of transitive actions from still images. Such actions are typically
based on an interaction between a body part and an action-object (the
object involved in the action). In a cognitive study of 101 frequent
actions and 15 body parts, Hidaka, Shohei and Smith \cite{maouene2008body}
have shown that hands and faces (in particular mouth and eyes) were
by far the most frequent body parts involved in actions. It is interesting
to note that in the human brain, actions related to hands and faces
appear to play a special role: neurons have been identified in both
physiological and fMRI studies \cite{brozzoli2009peripersonal,ladavas1998visual}
which appear to represent 'hand space' and 'face space', the regions
of space surrounding the hands and face respectively. In this work
we focus, but are not limited to such actions, in particular interactions
between the mouth and small objects, including drinking, smoking,
brushing teeth and others. We refer to them as face-related actions.
For such actions, the ability to detect the action-object involved
has a crucial effect on the attainable classification accuracy. A
simple demonstration is summarized in table \ref{tab:Classification-performance-impro}
(Section \ref{sec:The-Importance-of} below). The action objects involving
such classes are often barely visible, making them hard to detect
(see Figure \ref{fig:Actions-with-small}). Furthermore, the spatial
configuration of the action object with respect to the relevant body
parts must be considered to avoid the chance of misclassification
due to the detection of an unrelated action object in the image. In
other words, the right action-object should be detected at a particular
surrounding context.

To obtain these goals, our algorithm includes a method for the detection
of the relevant action object and related body parts as means to aid
the classification process. We seek the relevant body parts and the
action object \textit{jointly} using a cascade of two fully convolutional
networks: the first operates at the entire image level, highlighting
regions of high probability for the location of the action object.
The second network, guided by the result of the first, refines the
probability estimates within these regions. The results of both networks
are combined to generate a set of action-object candidates which are
pruned using contextual features. In the final stage, each candidate,
together with features extracted from the entire image and from the
two networks, is used to score the image as depicting a target action
class. We validate our approach on a subset of the Stanford-40 Actions
\cite{ICCV11_0744} dataset, achieving a 35\% improvement over state
of the art.

\subsection{Related Work\label{sub:Related-Work}}

Recent work on action recognition in still images can be categorized
into several approaches. One group of methods attempt to produce accurate
human pose estimation and then utilize it to extract features in relevant
locations, such as near the head or hands, or below the torso \cite{ICCV11_0744,desai2012detecting}

Others attempt to find relevant image regions in a semi-supervised
manner, given the image label. Prest et al \cite{eth_biwi_00828}
find candidate regions for action-objects and optimize a cost function
which seeks agreement between the appearance of the action objects
for each class as well as their location relative to the person. In
\cite{sener2012recognizing} the objectness \cite{alexe2010object}
measure is applied to detect many candidate regions in each image,
after which multiple-instance-learning is utilized to give more weight
to the informative ones. Their method does not explicitly find regions
containing action objects, but any region which is informative with
respect to the target action. In \cite{yao2011combining} a random
forest is trained by choosing the most discriminative rectangular
image region (from a large set of randomly generated candidates) at
each node, where the images are aligned so the face is in a known
location. This has the advantage of spatially interpretable results. 

Some methods seek the action objects more explicitly: \cite{ICCV11_0744}
apply object detectors from Object-Bank \cite{li2010object} and use
their output among other cues to classify the images. In \cite{DBLP:conf/icmcs/LiangWHL14},
outputs of stronger object detectors are combined with a pose estimation
of the upper body in a neural net-setting and show improved results,
where the object detectors are the main cause for the improvement
in performance. Very recently \cite{gkioxari2015contextual} introduced
R{*}CNN which is a semi-supervised approach to find the most informative
image region constrained to overlap with the person bounding box.
Also relevant to our work are some recent attention based methods:
\cite{xu2015show} uses a recurrent neural network which restricts
its attention to a relevant subset of features from the image at a
given timestep. We do not use an RNN, but two different networks.
Instead of restricting the view of the second network to a subset
of the original features, we allow the network to extract new ones,
allowing it to focus on more subtle details. 

Unlike \cite{eth_biwi_00828,sener2012recognizing,yao2011combining,gkioxari2015contextual},
our approach is fully supervised. Unlike\cite{DBLP:conf/icmcs/LiangWHL14}
we seek the relevant body parts and action objects jointly instead
of using separate stages. In contrast with \cite{gkioxari2015contextual},
we employ a coarse-to-fine approach which improves on the ability
of a single stage to localize the relevant image region accurately. 

The rest of the paper is organized as follows: in the next section,
we demonstrate our claim that exact detection of the action object
can be crucial for the task of classification. In Section \ref{sec:Approach},
we describe our approach in detail. In Section \ref{sec:Experiments}
we report on experiments used to validate our approach on a subset
of the Stanford-40 Actions \cite{ICCV11_0744} dataset. Section \ref{sec:Conclusions-=000026-Future}
contains a discussion \& concluding remarks.

\section{The Importance of Action objects in Action Classification\label{sec:The-Importance-of}}

\begin{figure}
\includegraphics[width=1\columnwidth]{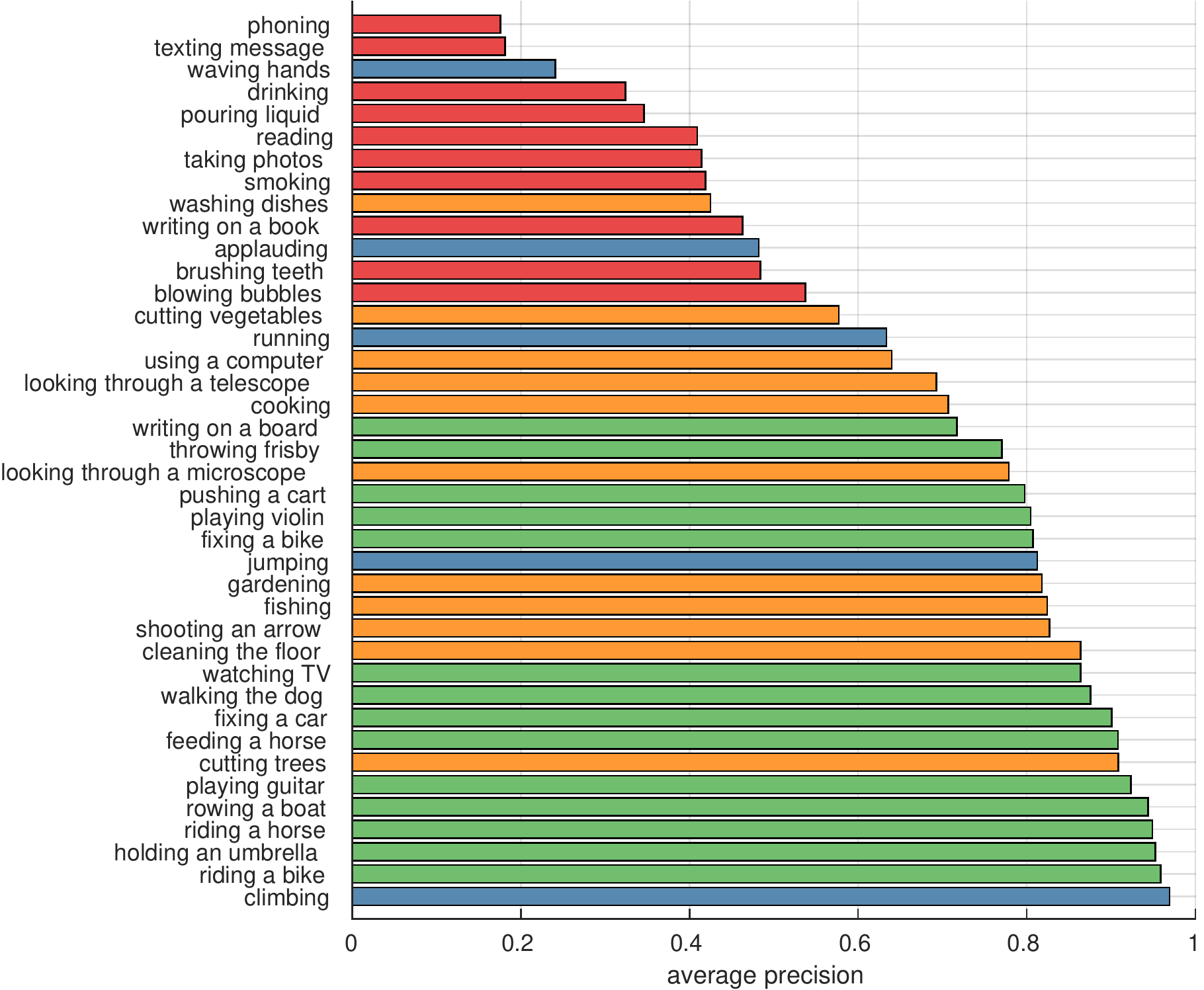}\protect\caption{\label{fig:Distribution-of-baseline}Distribution of baseline performance
for various action classes. We note the typical size of the action-objects
for the transitive actions (those involving objects), and classify
them as small(\textcolor{red}{red}), large(\textcolor{green}{green}),
and leave the rest of the classes undecided (orange). Mean performance
is significantly worse on actions involving small objects (37.5 \%)
vs. large objects (87 \%).}
\end{figure}

In this section, we demonstrate our claim that exact localization
of action objects plays a critical role in classifying actions, in
particular when they are hard to detect, \eg, due to their size or
occlusion. We begin by a simple baseline experiment on the Stanford-40
Actions \cite{ICCV11_0744} dataset. It contains 9532 images of 40
action classes split to 4000 for training and the rest for testing.
We train a linear SVM to discriminate between each of the 40 action
classes in a one-vs-all manner, using features from the penultimate
layer vgg-16 \cite{Simonyan2014}, \ie, fc6, and report performance
in terms of average precision. Figure \ref{fig:Distribution-of-baseline}
shows the resulting per-class performance. We have marked classes
where the action object tends to be very small or very large related
to the image. The difference in mean performance is striking: while
for the small objects, the mean AP is \textbf{37.5\%} and for the
large ones \textbf{87\%}. The mean performance on the entire dataset
is 67\%. Hence, we further restrict our analysis to a subset of Stanford-40
Actions \cite{ICCV11_0744} containing 5 classes: drinking, smoking,
blowing bubbles, brushing teeth and phoning - where the action objects
involved tend to be quite small. We name this the FRA dataset standing
for Face Related Actions. We augment the annotation of the images
in FRA by adding the exact delineation of the action objects and faces
and the bounding boxes of the hands. Next, we test the ability of
a classifier to tell apart these 5 action classes when it is given
various cues. We do this by extracting features for each image from
the following regions:
\begin{itemize}
\item Entire image (\textbf{G}lobal appearance)
\item Extended face bounding box scaled to 2x the original size (\textbf{F}ace
appearance)
\item Bounding box of action object (\textbf{O}bject appearance)
\item Exact region of action object ( \textbf{MO} : \textbf{M}asked \textbf{O}bject
appearance)
\end{itemize}
We extract fc6 features from each of these regions. For MO, the masked
object region, the image is cropped around the region bounding box
and all pixels not belonging to the object mask are set to the value
of the mean image learned by vgg-16. A linear SVM is trained on the
feature representation of each feature alone and on concatenations
of various combinations. We allow access to both face bounding boxes
and object bounding boxes/masks at test time, as if having access
to an Oracle which reveals them. The performance of the classifier
with different feature combinations is summarized in Table \ref{tab:Classification-performance-impro}.
Clearly, while the global (G) and face features (F) hold some information,
they are significantly outperformed by the classifier when it is given
the bounding box of the action object (O). This is further enhanced
by masking the surroundings of the action object (MO), which performs
best; The masking holds some information about the shape of the object
in addition to its appearance. Combining all features provides a further
boost, owing to the combination of local and contextual cues. 

\noindent In the next section, we show our approach to detecting the
action object automatically.
\begin{table}
\begin{tabular}{|c|c|c|c|c|c|c|}
\hline 
Region & \multirow{1}{*}{Mean AP} & \includegraphics[width=0.08\columnwidth]{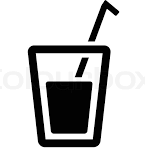} & \includegraphics[width=0.08\columnwidth]{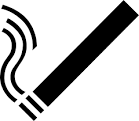} & \includegraphics[width=0.08\columnwidth]{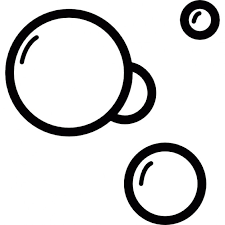} & \includegraphics[width=0.08\columnwidth]{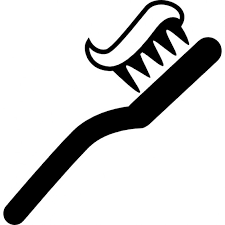} & \includegraphics[width=0.08\columnwidth]{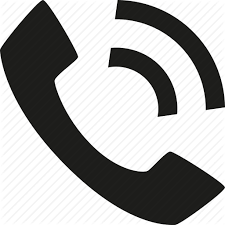}\tabularnewline
\hline 
\hline 
G & .58 & .54 & .58 & .74 & .57 & .45\tabularnewline
\hline 
Face & .69 & .61 & .57 & .82 & .72 & .76\tabularnewline
\hline 
O & .76 & .87 & .80 & .79 & .55 & .81\tabularnewline
\hline 
MO & .82 & .91 & .78 & .85 & .66 & .88\tabularnewline
\hline 
G+Face+O & .88 & .90 & .92 & .91 & .80 & .90\tabularnewline
\hline 
All & .91 & .94 & .93 & .93 & .85 & .90\tabularnewline
\hline 
\end{tabular}

\protect\caption{\label{tab:Classification-performance-impro}Classification performance
improves drastically as classifier is given access to the action objects'
exact locations : \textbf{G}lobal, Face, action \textbf{O}bject bounding
box, \textbf{MO}: \textbf{M}asked Object. This demonstrates the need
for accurately detecting the action-object.}
\end{table}

\section{Approach\label{sec:Approach}}

\begin{figure*}
\includegraphics[width=1\textwidth]{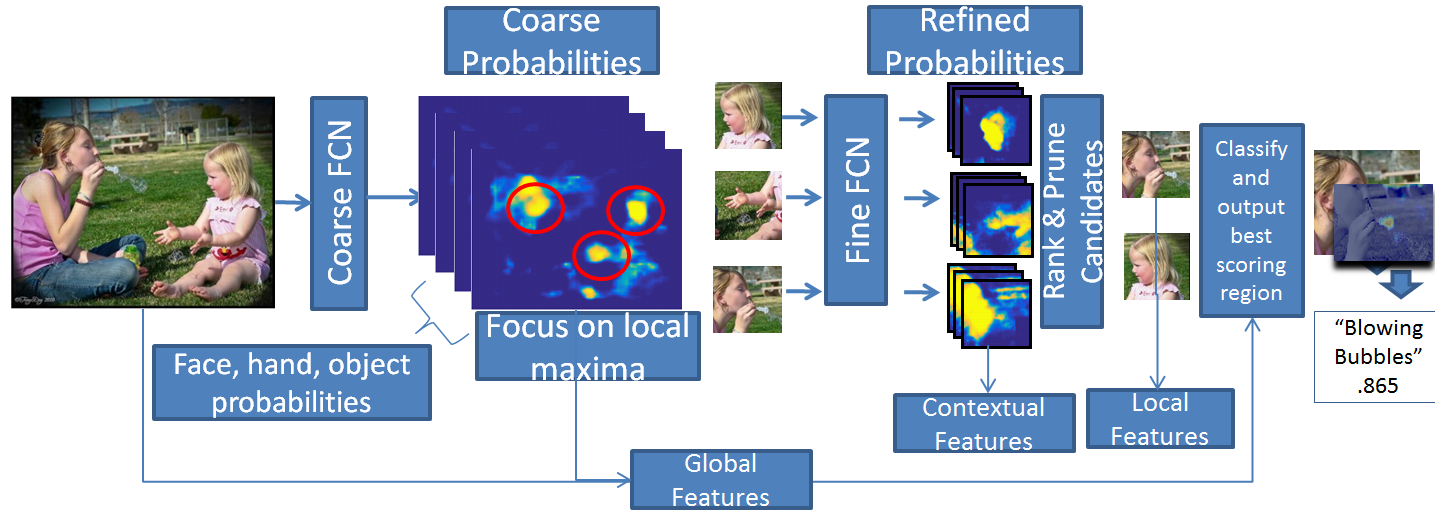}

\protect\caption{\label{fig:Flow-of-proposed}Flow of proposed method. A fully convolutional
network is applied to predict body parts and action objects. This
guides a secondary network to refine the predictions on a few selected
image regions. Using contextual features, the regions are ranked \&
pruned. The remaining candidates are scored using both global and
local features to produce a final classification, along with a visualization
of the detected action object. }
\end{figure*}

Our goal is to classify the action being performed in an image. To
that end, we perform the following stages: 

\textbullet{} Produce a probability estimate for the locations of
hands, head and action objects in the entire image
\begin{itemize}
\item Refine the probability estimates at a set of selected locations
\item Using the above probabilities, generate a set of candidate action
objects 
\item Rank and prune the set of candidate objects by using contextual features
\item Extract features from the probabilistic estimates and appearances
of the image and predicted action object locations to produce a final
classification. 
\end{itemize}
We now elaborate on each of these stages.

\subsection{Coarse to Fine Semantic Segmentation \label{sub:Coarse-to-Fine_3.1}}

We begin by producing a semantic segmentation of the entire image,
then refining it: we train a fully convolutional network as in \cite{long2014fully}
to predict a pixel-wise segmentation of the image. The network is
trained to predict \textbf{jointly} the location of faces, hands,
and $k$ different object categories relating to the different action
classes. Using the framework of \cite{vedaldi15matconvnet}, we do
so by fine-tuning the vgg-16 network of \cite{Simonyan2014}. We denote
the set of learned labels as $\mathcal{L}$: 

\begin{align}
\mbox{\ensuremath{\mathcal{L}}} & =\{bg,face,hand\}\cup Obj
\end{align}
where $Obj=\{obj_{1}\text{\dots}obj_{k}\}$ is the set objects relating
to $k$ action classes. We name this first network $Net_{coarse}$. 

For a network $N$ and image $I$, we denote by $P^{N}$ the probability
maps resulting from applying $N$ to $I$:

\begin{equation}
N(I)=P^{N}(I,x,y,c),c\in\mathcal{L}
\end{equation}

\noindent where the superscript $\cdot^{N}$ is used to indicate the
network which operated on the image. In other words, $N$ assigns
to each pixel in $I$ a probability for each of the classes. For brevity,
we might drop the parameters where it is appropriate, writing only
$P^{N}$. For clarity, we may also write $P^{coarse}$ instead of
$P^{\ensuremath{Net_{coarse}}}$.

The predictions of this $Net_{coarse}$, albeit quite informative,
can often miss the action object or misclassify it as belonging to
another category (see Section \ref{sec:Experiments}). To improve
the localization and accuracy of the estimate it produces, we train
a secondary network; we use the same base network (vgg-16) but on
a different extent whose purpose is to refine the predictions of the
first network. We name the second network $Net_{fine}$. It is trained
on sub-windows of the original images which are upscaled during the
training and testing process. Hence it is trained on a data of a different
scale than $Net_{coarse}$. Moreover, since it operates on enlarged
sub-windows each image, its output is eventually transformed to a
finer scale in the original image's coordinate system. We note that
in our experiments, attempting to train the 32-pixel stride version
of Long et al \cite{long2014fully} worked significantly less well
as the full 8-pixel stride version, which uses intermediate features
from the net's hierarchy to refine the final segmentation. Both networks
are trained using the full 8-pixel stride method. 

For each training image $I_{j}:j\in{1\ldots N}$, we define $B_{j}$
to be a sub-window of the entire image whose extent allows a good
tradeoff between the image region to be inspected (which is desirably
small to capture fine details), yet contains the desired objects or
body parts. The details of how this bounding box is selected depend
on the nature of the dataset and are described in Section \ref{sub:Training}.
The training set for $Net_{fine}$ is obtained by cropping each training
image $I_{j}$ around $B_{j}$ and cropping the ground-truth label-map
accordingly. 

$Net_{fine}$ is used to refine the output of $Net_{coarse}$ as follows:
after applying $Net_{coarse}$ to an image $I$ we seek the $m$ top
local maxima of $P^{Coarse}$ which are at least $p$ pixels apart.
A window is cropped around each local maxima with a fixed size relative
to the size of the original image and $Net_{fine}$ is applied to
that sub-window after proper resizing. The probabilities predicted
for overlapping sub-windows are averaged. We chose $m=5$, $p=20$
by validating on a small subset of the training set. Denote the resulting
refined probability map by $P{}^{fine}$. See Figure \ref{fig:Coarse_And_Fine}
for an example of the resultant coarse and fine probability maps and
the resulting predictions. 

\begin{center}
\begin{figure*}
\begin{centering}
\subfloat[Coarse prediction]{\begin{centering}
\includegraphics[height=0.22\textwidth]{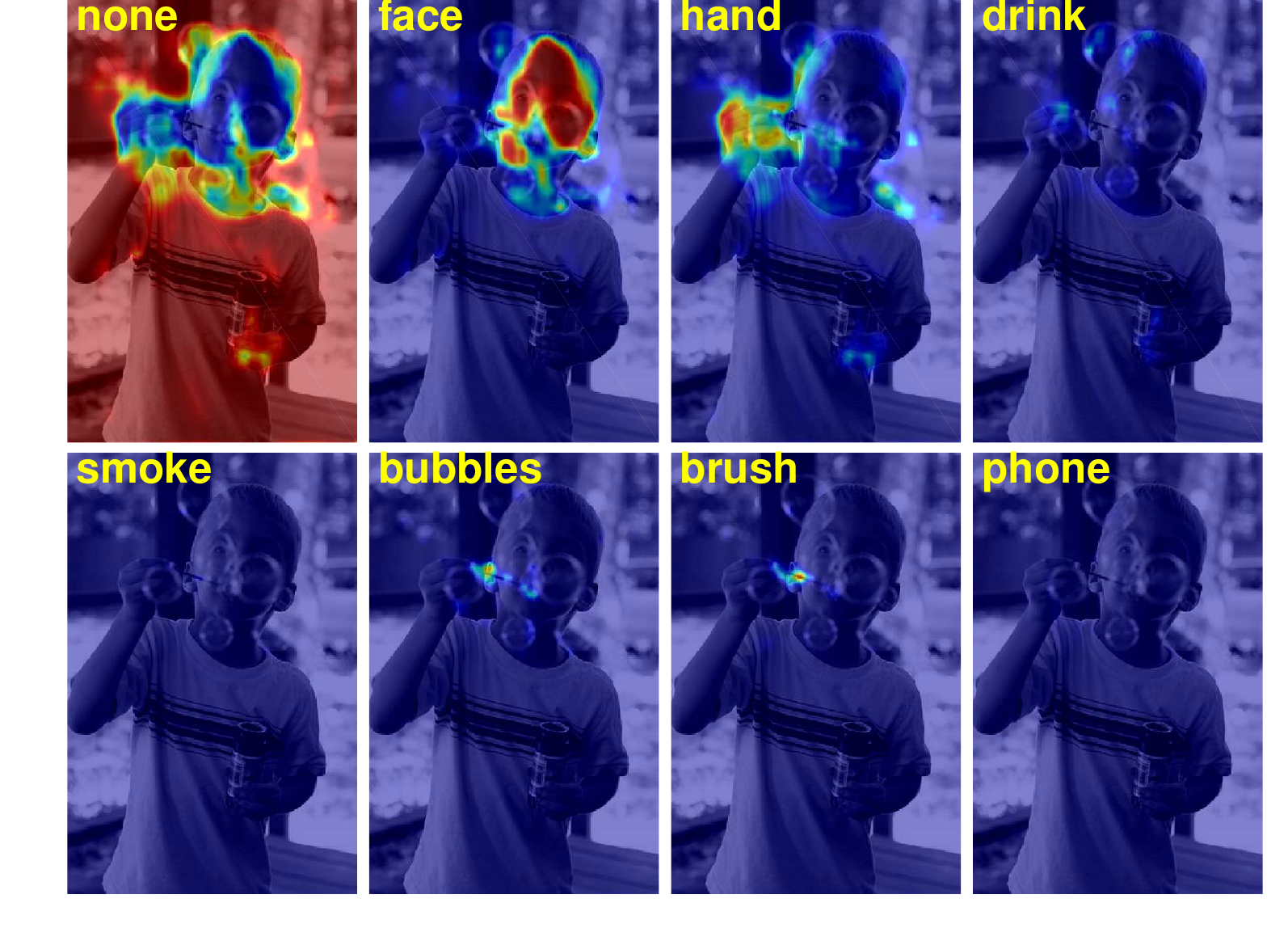}\includegraphics[bb=0bp 0bp 325bp 273bp,height=0.2\textwidth]{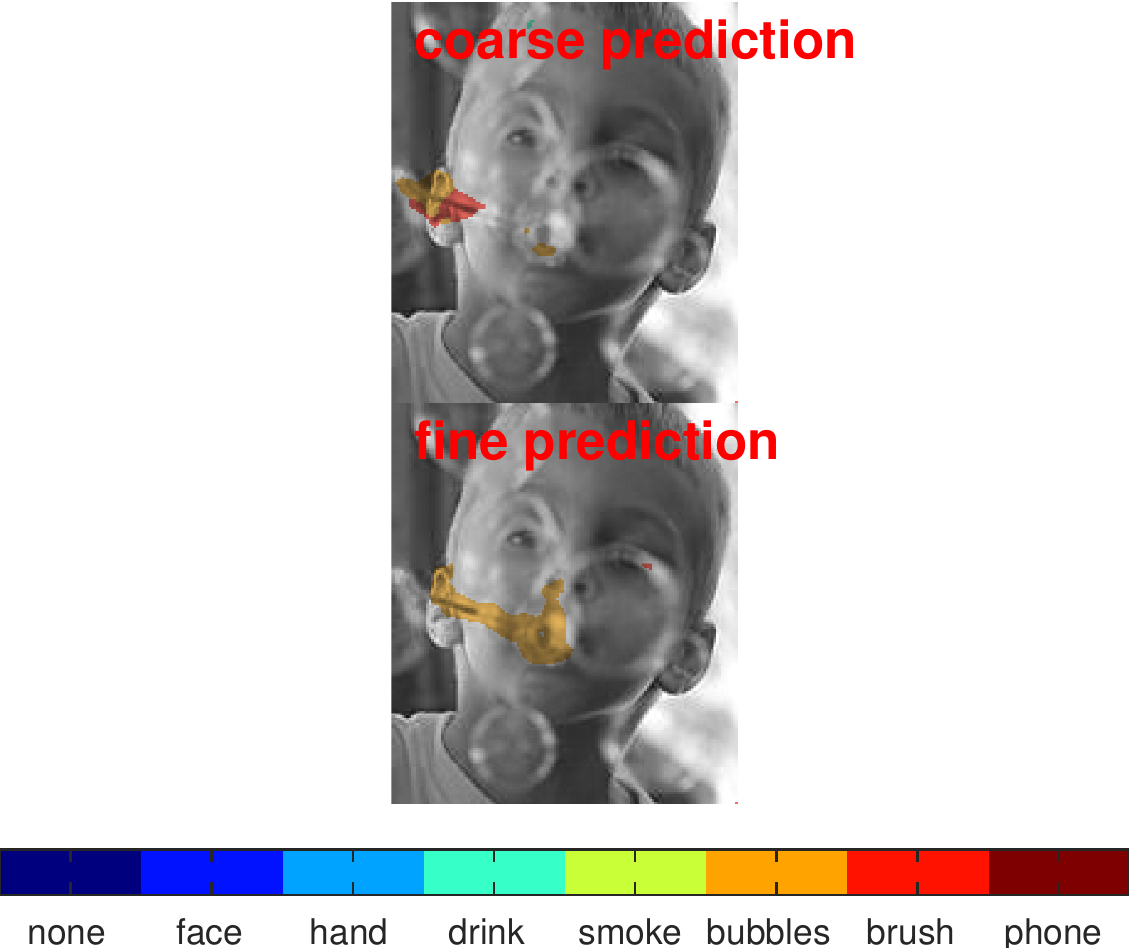}
\par\end{centering}

}\subfloat[Fine prediction]{\includegraphics[height=0.22\textwidth]{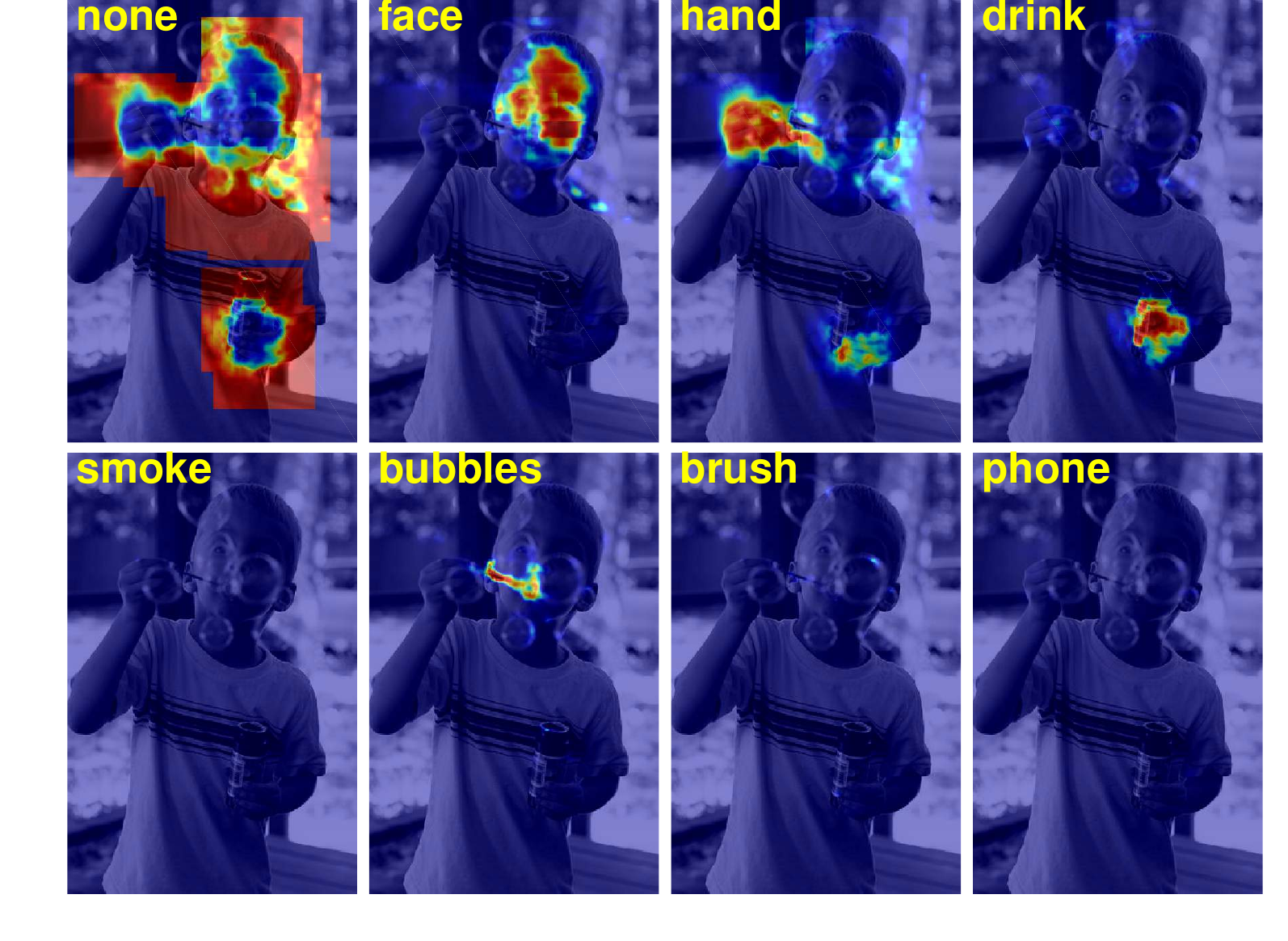}

}
\par\end{centering}

\protect\caption{\label{fig:Coarse_And_Fine}Coarse to fine predictions. (a) $Net_{coarse}$
is applied to the entire image, producing (left) a pixelwise per-class
probability map and (center-top) prediction; (b) guided by local maxima
of $Net_{coarse}$, \textbf{$Net_{fine}$} is applied to a selected
set of subwindows. A missed bubble wand is detected by the refinement
process, as well as other fine details. Probability of false classes
(\eg,brushing) are suppressed, see differences in predicted probability
maps. Best viewed in color online.}
\end{figure*}

\par\end{center}

\subsection{Contextual Features\label{sub:Contextual-Features}}

We now describe how we use the outputs for both networks to generate
candidate regions for action objects. Identifying the region corresponding
to the object depends on its on appearance as well as the context,
e.g. of the face and hand. The outputs of the coarse and fine networks
often produce good candidate regions, along with many false ones:
typically tens of regions per image. In addition, the location of
the candidates may be correct but their predicted identity (i.e, the
maximal probability) is wrong. Therefore we produce candidate regions
from the resultant probability maps in two complementary ways: Define
\begin{equation}
Pred^{N}(x,y)=argmax_{c\in Obj}P^{N}(x,y,c)
\end{equation}
 $Pred^{N}(x,y)$ is the pixelwise prediction of net $N\in\{Net_{coarse},Net_{fine}\}$.
We denote by $R_{pred}$ the set of connected components in $Pred^{N}$.
In addition, let $M_{p}$ be the set of $l$ local maxima, computed
separately for each channel $c\in Obj$ of $P^{N}$. For each channel
$c\in Obj$ we apply the adaptive thresholding technique of Otsu \cite{otsu1975threshold}.
We denote by $R_{m}$ the union of the regions obtained by the thresholding
process. We remove from $R_{m}$ regions which do not contain any
element in $M_{p}.$ Finally, we denote 
\begin{equation}
R=R_{pred}\cup R_{m}
\end{equation}

\noindent as the set of candidate regions for the image. $R_{m}$
and $R_{pred}$ are complementary since the maximal prediction of
the network may not necessarily contain a local maxima in any of the
probability channels. 

We train a regressor to score each candidate region: Let $R_{i}\in\{1...|R|\}$
be a candidate region and $B(R_{i})$ its bounding box. We extract
short and long range contextual features for $R_{i}$: for the \textit{\textcolor{black}{short}}
range features we scale $B(R_{i})$ by a factor of 3 while retaining
the box center. Next, we split the enlarged bounding box into a $5\times5$
grid, where we assign to each grid cell the mean value of each channel
of the probability maps inside that cell. Formally, let $w$ be the
window defined by $B(R_{i})$ and $w_{r,c}$,$r,c\in\{1\ldots5\}$
the subwindow for the $r,c$ row/column. We define

\begin{equation}
F_{context}^{s}(R_{i},r,c)=\sum_{x,y\in w_{r,c}}\frac{P(x,y,c)}{|w_{r,c}|}
\end{equation}
where $|w_{r,c}|$ is the area of each grid cell in pixels. This yields
a feature vector of length $5\times5\times\mathcal{\ensuremath{L}}$
channels representing the values of the network\textquoteright s predictions
both inside the predicted region and in the immediate surroundings.
Similarly, define a second bounding box, $B^{l}(R_{i})$ to be a bounding
box 1/3 times the size of the image with the same center as $B(R_{i})$.
We define a $5\times5$ grid inside $B^{l}(R_{i})$ and use this to
extract \textit{long} range contextual features $F_{context}^{L}$
in the same manner as for the short range $F_{context}^{s}.$ We compute
these features and concatenate them for all candidate regions from
the training set (including the ground truth ones). We train a regressor
whose input is these features and the target output is the overlap
with the bounding boxes of ground-truth regions. 

It is worth noting that we have attempted to incorporate these contextual
features as another layer in the network, as they effectively serve
as another pooling layer over the probability maps. Perhaps due to
the small training set we have used, this network did not converge
to results as good as those obtained by our context features. At least
for a dataset of this size, constraining the context features to be
of this form performed better than letting the network learn how to
perform the pooling on its own. It remains an open question if a larger
dataset would facilitate end-to-end learning, effectively learning
the contextual features as an integral part of the network.

\subsection{Classification}

We now show how all the features are combined into a final classification.
We extract the features as follows: for a network $N_{t}\in\{Net_{coarse},Net_{fine}\}$
we denote

\begin{equation}
F^{t}(I)=[S_{1}^{N_{t}},S_{2}^{N_{t}},\dots,S_{L}^{N_{t}}]^{T}\label{eq:global_score}
\end{equation}

\noindent where $S^{t}(I)_{c}=max_{x,y}\{P^{t}(x,y,c)\}$. $F^{t}(I)$
is a concatenation of the maximal values of each channel, including
those predicting the body parts; it serves us as a global representation
of the image. Note that while features extracted this way are significantly
more discriminative when $N_{t}=Net_{fine},$we found that the networks
are complementary; combining $F^{t}(I)$ from both networks works
better than each own its own. Next we add features from the action
object and face regions. The face is detected using the face detector
of \cite{mathias2014face}. We increase the size of each image by
2 before running the face detector and retain the single highest scoring
face per image. Using the regressor learned in Section \ref{sub:Contextual-Features},
we rank the candidate regions $R{}_{i}$ and retain the top $q$ regions
per image. For each region, we extract appearance features in the
form of fc6 features using the vgg-16 {[}18{]} network. In training,
we use the ground-truth regions. Let $F_{f}(I)$, $F_{obj,i}(I)$
be the feature representations for the face and $q=3$ candidate object
features respectively, where $i=1\dots q$ are the indices of the
top-ranked regions. Let $F_{g}(I)$ be the fc6 representation of the
entire image. We extract features from the candidate regions by using
both bounding boxes and masked versions as in section \ref{sec:The-Importance-of},
with candidate regions $R_{i}$ as masks. The final scoring of image
$I$ for class $c$ is defined as: 

\begin{equation}
S(I)_{c}=max_{i}W_{c}^{T}\cdot F_{i}(I)\label{eq:final_score}
\end{equation}
 where $F_{i}(I)=[F_{g},F^{coarse},F^{fine},F_{f},F_{obj,i}]^{T}$
(dropping the $I$ argument for brevity) and $W_{c}^{T}$ is the set
of weights learned by an SVM classifier in a one-versus all manner
per action class. Please refer to section \ref{sec:Experiments} for
experiments validating our approach.

\subsection{Training\label{sub:Training}}

To train $Net_{coarse}$, we construct a fully-convolutional neural
net with the DAG architecture\cite{long2014fully} to produce a prediction
of 8-pixel stride by fine-tuning the vgg-16 network. We use a learning
rate of .0001 for 100 epochs, as our training set contains only a
few hundreds of images. A similar procedure is done for $Net_{fine}$,
where the training samples are selected as described in Section \ref{sub:Coarse-to-Fine_3.1}.
For the FRA dataset, the sub-windows are selected to be the ground-truth
bounding box of each face scaled by a factor of 2, this includes most
of the action objects. Note that for FRA we did not use the provided
bounding boxes at test time. The regressor for predicting the locations
of action objects is a support-vector regression trained on the context
features around each candidate region (see Section \ref{sub:Contextual-Features}).
All classifiers are trained using an SVM with a regularization $\lambda=.001$.

\section{Experiments\label{sec:Experiments}}

\begin{center}
\begin{table*}[t]
\begin{centering}
\begin{tabular}{|l|l|l|l|l|l|l|l|l|l|l|}
\hline 
\textbf{G} & \textbf{Face} & \textbf{C} & \textbf{F} & \textbf{Obj} & \multicolumn{1}{l}{\textbf{Mean AP}} & \includegraphics[width=0.08\columnwidth]{figures/icons/drink} & \includegraphics[width=0.08\columnwidth]{figures/icons/cigarette} & \includegraphics[width=0.08\columnwidth]{figures/icons/bubbles} & \includegraphics[width=0.08\columnwidth]{figures/icons/toothbrush} & \includegraphics[width=0.08\columnwidth]{figures/icons/phone}\tabularnewline
\hline 
 & + & + & + & + & 0.845 & 0.840 & 0.889 & 0.913 & 0.743 & 0.839\tabularnewline
\hline 
+ &  & + & + & + & 0.851 & 0.836 & 0.907 & 0.909 & 0.751 & 0.851\tabularnewline
\hline 
+ & + &  & + & + & 0.856 & 0.842 & 0.898 & 0.907 & 0.771 & 0.865\tabularnewline
\hline 
+ & + & + &  & + & 0.830 & 0.818 & 0.841 & 0.905 & 0.753 & 0.835\tabularnewline
\hline 
+ & + & + & + &  & 0.848 & 0.801 & 0.895 & 0.905 & 0.780 & 0.860\tabularnewline
\hline 
+ & + & + & + & + & \textbf{0.865} & \textbf{0.845} & \textbf{0.910} & \textbf{0.914} & \textbf{0.786} & \textbf{0.868}\tabularnewline
\hline 
\end{tabular}
\par\end{centering}

\protect\caption{\label{tab:Ablation-Study}Ablation Study of various features combinations.
\textbf{G}lobal, \textbf{Face}, \textbf{F}ine/\textbf{C}oarse segmentation,
action \textbf{Obj}ect features). Removing the fine phase has the
worst effect on performance. }
\end{table*}

\par\end{center}

\begin{table}
\begin{tabular}{|l|l|l|l|l|l|l|}
\hline 
 & \textbf{MAP} & \includegraphics[width=0.08\columnwidth]{figures/icons/drink} & \includegraphics[width=0.08\columnwidth]{figures/icons/cigarette} & \includegraphics[width=0.08\columnwidth]{figures/icons/bubbles} & \includegraphics[width=0.08\columnwidth]{figures/icons/toothbrush} & \includegraphics[width=0.08\columnwidth]{figures/icons/phone}\tabularnewline
\hline 
C & 0.636 & 0.593 & 0.692 & 0.720 & 0.551 & 0.623\tabularnewline
\hline 
G & 0.642 & 0.601 & 0.623 & 0.789 & 0.660 & 0.537\tabularnewline
\hline 
F & 0.668 & 0.565 & \textbf{0.816} & 0.697 & 0.513 & \textbf{0.749}\tabularnewline
\hline 
Face & 0.699 & 0.658 & 0.578 & 0.829 & \textbf{0.686} & 0.742\tabularnewline
\hline 
Obj & \textbf{0.743} & \textbf{0.776} & 0.744 & \textbf{0.851} & 0.606 & 0.738\tabularnewline
\hline 
\end{tabular}

\protect\caption{\label{tab:Single-feature-performance}Single feature performance
for classifying actions. (\textbf{G}lobal, \textbf{Face}, \textbf{F}ine/\textbf{C}oarse
segmentation,action\textbf{ Obj}ect features) On average, features
extracted from the automatically detected action-object outperform
others.}
\end{table}

\begin{figure}
\subfloat[drinking]{\includegraphics[width=1\columnwidth]{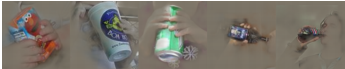}

}

\subfloat[smoking]{\includegraphics[width=1\columnwidth]{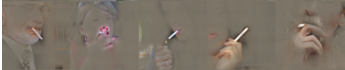}}

\subfloat[blowing bubbles]{\includegraphics[width=1\columnwidth]{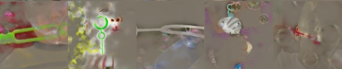}}

\subfloat[brushing teeth]{\includegraphics[width=1\columnwidth]{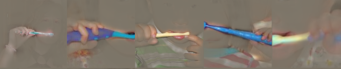}

}

\subfloat[phoning]{\includegraphics[width=1\columnwidth]{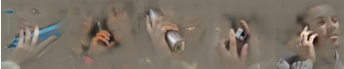}}\protect\caption{\label{fig:Highly-scoring-detected}Highly scoring action-objects
detected by our method, weighted by the action-object per pixel probability
for the respective class.}
\end{figure}

We now show some experimental results. We begin with FRA, a subset
of Stanford-40 Actions dataset, containing 5 action classes: drinking,
smoking, blowing bubbles, brushing teeth and phoning. To show the
contribution of each feature, we perform the following ablation study.
First, we test the performance of each feature on its own. In Table
\ref{tab:Single-feature-performance}, we can see that while the coarse
level  information extracted by $Net_{coarse}$ provides moderate
results, it is outperformed by other feature types, namely the global
image representation and that of the face. The fine representation
performs on average slightly less well as the extended face area,
with the biggest exception being the smoking category, where it improves
from an AP of .578 to .816 (in many cases the cigarettes are held
far away from the face). As can be seen in the images of Figure \ref{fig:Highly-scoring-detected},
the semantic  segmentation of \textbf{$Net_{fine}$ }is able to capture
well the cigarettes in the image. The \textbf{Obj} score was obtained
by assigning to each image that of the highest scoring candidate region.
We can see that its performance nears that of the ``Oracle'' classifier,
which is given the bounding box at test time. See Figure \ref{fig:Highly-scoring-detected}
for an example of top-ranked detected action objectes weighted by
the predicted object probabilities. 

Next, we performed an ablation study showing how performance changes
when we use all of the sources of information except one. Table \ref{tab:Ablation-Study}
summarizes this. We can see that performance is worst when excluding
the $Net_{fine}$ predictions. Also, as expected by our motivation
of the problem in Section \ref{sec:The-Importance-of}, we see that
the best performance is gained when including the predicted object
locations. Overall, the increase in mean average precision obtained
via the baseline global features produced by the vgg-16 network increases
from 0.642 to .865, a 35\% relative increase. Note that this is also
quite near the results of the ``Oracle'' classifier (Section \ref{sec:The-Importance-of}).

\subsection{Joint training for Pose and Objects}

It is noteworthy that the joint training of the networks to detect
the hands and faces along with the action objects performed dramatically
better than attempting to train a network to predict the location
of the action objects only; we have attempted to train a network when
the ground-truth masks contained only the locations and identities
of action objects without body parts. This worked very poorly; we
conjecture that while the action objects may be difficult to detect
on their own, contextual cues are implicitly learned by the networks.
Such cues likely include proximity to the face or hand.

\section{Conclusions \& Future Work\label{sec:Conclusions-=000026-Future}}

We have demonstrated that exact localization of action objects and
their relation to the body parts is important for action classes where
the action-objects are small and often barely visible. We have done
so using two main elements. First, using a coarse-to-fine approach
which focuses in a second stage on relevant image regions. Second,
using contextual cues which aid the detection of the small objects.
This happens both during the network's operation, as it seeks the
objects jointly with the relevant body parts, as well as pruning false
object candidates generated by the networks, by considering their
context explicitly. The coarse to fine approach whos networks are
based on the full 8-pixel stride model of \cite{long2014fully} utilizes
features from intermediate levels of the network and not only from
the top level. It outperforms a purely feed-forward method such as
obtained from the 32-pixel stride version. Together, these elements
aid in good localizations of action objects, leading to a significant
improvement over baseline methods. Our method uses two networks and
a specific form of contextual features. Our comparisons showed that
the results are better than incorporating the entire process in an
end-to-end pipeline; it remains an open question if this is due to
the relatively small size of the training set. A current drawback
of the method is that it required annotation of both body-parts and
action objects in the dataset; in the future we intend to alleviate
this constraint by being able to combine information from existing
datasets, which typically contain annotations of objects or poses,
but not both.

\bibliographystyle{plain}
\bibliography{bibtex_db}

\end{document}